\documentclass[runningheads]{llncs}
\usepackage{graphicx}
\usepackage{xcolor}
\usepackage{booktabs}
\begin{document}
\title{Decontextualized I3D ConvNet for ultra-distance runners performance analysis at a glance\thanks{Supported by the ULPGC under project \mbox{ULPGC2018-08}, by the Spanish Ministry of Economy and Competitiveness (MINECO) under project RTI2018-093337-B-I00, by the Spanish Ministry of Science and Innovation under project PID2019-107228RB-I00, and by the Gobierno de Canarias and FEDER funds under project ProID2020010024.}}
\titlerunning{Decontextualized I3D ConvNet for runners performance analysis at a glance}
\author{David Freire-Obreg\'on\inst{1}\orcidID{0000-0003-2378-4277} \and
Javier Lorenzo-Navarro\inst{1}\orcidID{0000-0002-2834-2067} \and
Modesto Castrill\'on-Santana\inst{1}\orcidID{0000-0002-8673-2725}}

%\author{First Author\inst{1}\orcidID{0000-1111-2222-3333} \and
%Second Author\inst{2,3}\orcidID{1111-2222-3333-4444} \and
%Third Author\inst{3}\orcidID{2222--3333-4444-5555}}
%
\authorrunning{David Freire-Obreg\'on et al.}
\institute{SIANI\\
Universidad de Las Palmas de Gran Canaria\\
Spain\\
\email{david.freire@ulpgc.es}}
%\institute{Princeton University, Princeton NJ 08544, USA \and
%Springer Heidelberg, Tiergartenstr. 17, 69121 Heidelberg, Germany

\maketitle              % typeset the header of the contribution
\begin{abstract}
In May 2021, the site runnersworld.com published that participation in ultra-distance races has increased by 1,676\% in the last 23 years. Moreover, nearly 41\% of those runners participate in more than one race per year. The development of wearable devices has undoubtedly contributed to motivating participants by providing performance measures in real-time. However, we believe there is room for improvement, particularly from the organizers point of view. This work aims to determine how the runners performance can be quantified and predicted by considering a non-invasive technique focusing on the ultra-running scenario. In this sense, participants are captured when they pass through a set of locations placed along the race track. Each footage is considered an input to an I3D ConvNet to extract the participant's running gait in our work. Furthermore, weather and illumination capture conditions or occlusions may affect these footages due to the race staff and other runners. To address this challenging task, we have tracked and codified the participant's running gait at some RPs and removed the context intending to ensure a runner-of-interest proper evaluation. 
The evaluation suggests that the features extracted by an I3D ConvNet provide enough information to estimate the participant's performance along the different race tracks.
\end{abstract}

\section{Introduction}
\label{sec:intro}

Our ability to evaluate an athlete's performance depends on the sporting context. For instance, in a soccer game scenario, the players gait/pose or precision when kicking the ball may provide valuable insights about the players condition. Similarly, the way an ultra-distance participant is running (i.e., gait, pace, and trajectory) may also provide some intuitions about the runner performance. However, despite numerous potential applications and existing wearable technologies, this ability remains a challenge for state-of-the-art visual recognition systems.

In contrast to the purpose of traditional human action recognition (HAR) to infer the label from predefined action categories~\cite{Tran18}, the aim of human action evaluation (HAE) is to automatically quantify how well people perform actions given a particular metric. HAE has been mainly exploited in applications for healthcare and rehabilitation~\cite{Parmar16}, self-learning platforms for practicing professional skills, or sports~\cite{Ghasemzadeh11}. In this regard, different approaches have been considered to tackle this problem in the sporting context, i.e., virtual reality~\cite{Bideau10} and wearable sensors~\cite{Tedesco20}. The former main drawback is the indoor environment restriction. In contrast, the latter can be considered an invasive proposal requiring a device calibration and may lead to some privacy concerns.

\begin{figure}[t]  
% \begin{minipage}[t]{1\linewidth}
    \centering
    \includegraphics[scale=0.55]{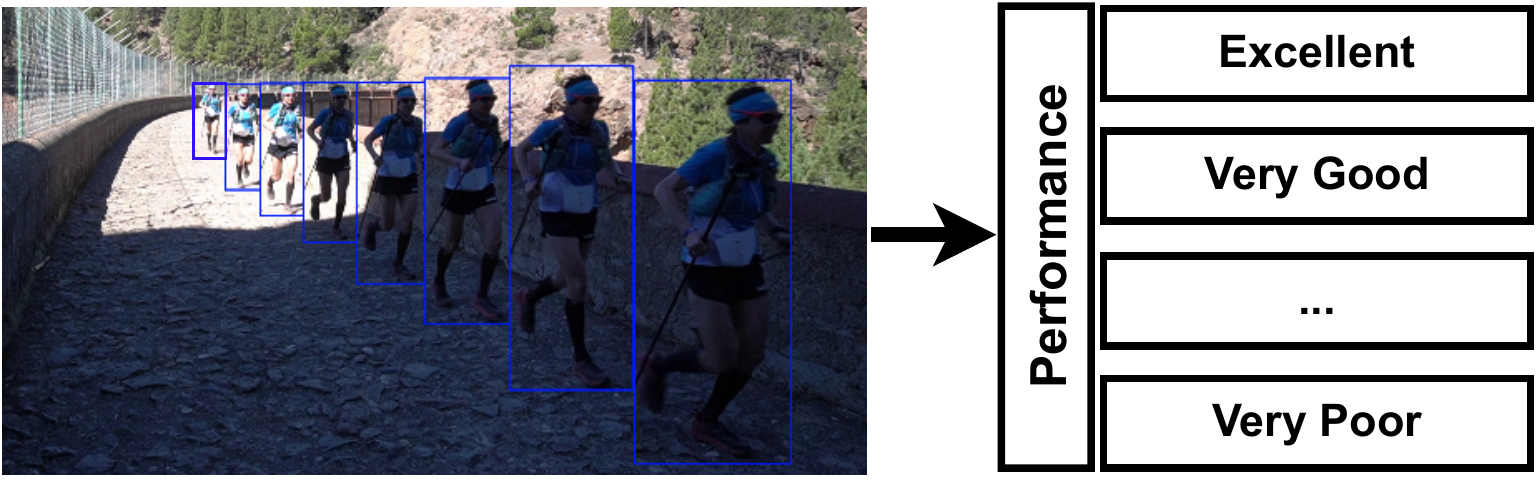}
    %   \end{minipage}
    \caption{\textbf{Runner video sequence.} We evaluate the runner performance with a model that efficiently classifies him/her motion into several categories in a single forward pass. Each category quantifies the runner's performance. The Figure shows frames of a clip present in the considered dataset \cite{Penate20-prl}.
     \label{fig:performance}}
\end{figure}

In this work, we take a step towards the athlete performance evaluation by processing short video clips recorded at different recording points (RPs) of an ultra-running track as input. First, we introduce a performance classifier built on top of a pre-trained deep neural network that reports highly competitive results in HAR. As can be seen in Figure \ref{fig:performance}, the classifier provides an output that represents the athlete performance. The performance is discretized by categorizing the runners RP qualification time into a set of categories (i.e., excellent, very good, and so on). 
We have conducted several experiments to predict the runner performance, considering the runner video clip as input. Moreover, we have evaluated our model to predict the current RP performance given a video clip and the following RP performance estimation given the current RP video input. Additionally, we have developed a thoughtful context analysis to show the relevance of the environment in the proposed pipeline. 

This work aims at answering some interesting questions: Can the runner performance be estimated from his/her motion? Is the context relevant? To what extent can the context be removed? Does this simplification come with a cost? 
We have evaluated our model in a dataset collected to evaluate Re-ID methods in complex real-world scenarios. The dataset contains $214$ ultra-distance runners captured at different RPs along the track. The achieved results are remarkable (up to an $83.7$\% of accuracy), and they have also provided interesting insights. The first one is that increasing the number of categories for quality assessment negatively affects, as expected, the classifier performance. Another insight is related to the importance of contextual information for the pre-trained I3D ConvNet and the limitations observed during the transfer learning.

The paper is organized into five sections. The next section discusses some related work. Section \ref{sec:pipeline} describes the proposed pipeline. Section \ref{sec:results} reports the experimental setup and the experimental results. Finally, conclusions are drawn in Section \ref{sec:con}.

%Recently, during the Euro 2020, the danish playmaker  Eriksen needed life-saving cardiac massage from medical team after collapsing. Seconds before collapsing, Eriksen went to collect the ball but fell limp, the ball bouncing off his knee as he slumped to the grass.

\section{Related Work}
\label{sec:sota}

Contrary to traditional HAR systems~\cite{Tran18,Varol18}, HAE systems~\cite{Patrona18} are designed not to identify a specific action but to capture the motion of the human body and to measure the completion quality of the captured motion through an evaluation technique. Both systems share common properties in terms of significant stages: First, a human detection algorithm locates the region-of-interest (ROI) in the image~\cite{Tu19}. Next, a pose representation algorithm~\cite{Cao18,He20} is computed all over the ROI to extract a set of keypoints representing the human pose. In this regard, the skeleton-based representation approach has been widely used as a data source to solve the human pose representation~\cite{Patrona18}. Finally, a deep neural network~\cite{Ye19} is trained on the extracted features representation for classification/regression purposes. This same primary approach has also been applied to some lower abstraction-level problems such as gait~\cite{Freire20} or hand-action~\cite{Simon17} recognition systems.

Athletes can measure their motion quality by judging their postures and movements through HAE techniques. The human pose representation plays a crucial role in evaluating the performed action. Lei et al. have identified three primary pose representations for the human action quality assessment~\cite{Lei19}. The challenge relies on finding robust features from pose sequences and establishing a method to measure the similarity of pose features~\cite{Wnuk10}. Second, the aforementioned skeleton-based representations encode the relationship between joints and body parts~\cite{Paiement14,Patrona18}. However, the estimated skeleton data can often be noisy in realistic scenes due to occlusions or changing lighting conditions~\cite{Carissimi18}. Finally, the deep learning methods for assessing the athlete's action quality.
In this representation approach, convolutional neural networks (CNN) can be combined with recurrent neural networks (RNN) due to the temporal dimension provided by the video input~\cite{Parmar17}. A typical network used for sports quality assessment is the 3D convolution network (C3D). This deep neural network that learns spatio-temporal features is increasingly being used for HAR~\cite{Varol18}. Precisely, our work can be framed in the deep learning methods for assessing the athlete's action quality suggested by Lei et al. We make use of the Inflated 3D ConvNet (I3D), which has been used to tackle the HAR problem in the past~\cite{Carreira17,Freire22}. This network passes a two-stream input (RGB and flow) through a combination of 3D convolutions, Inception modules, and max-pooling layers. It uses asymmetrical filters for max-pooling, maintaining time while pooling over the spatial dimension.

Several sports datasets have been collected in the past few years. Most of them were collected from international competitions events. In this regard, some of the most notable datasets are MTL-AQA~\cite{Parmar19}, UNLV AQA-7~\cite{Parmar19wacv} and Fis-V~\cite{Xu20}. The sports collected in those datasets are usually practiced indoors or in a not-occlusive environment, i.e.,  diving, skating, skiing, snowboarding, and trampoline. Also, the sports exhibitions in those datasets take no longer than a few seconds or minutes. Our work considers an ultra-distance race collection where professional and non-professional runners compete in a 30 hours race. There is a high set of variations in terms of lighting conditions, backgrounds, occlusions, and accessories due to the duration of the race. We strongly believe that our non-invasive quality assessment can provide relevant cues about the runner performance variation.

In summary, the work presented in this paper evaluates the race participant's performance considering an I3D network. The considered dataset provides a scenario in the wild. Further, we evaluate our pipeline, considering the raw video sequence as input and segmenting the runners to analyze the importance of context in I3D networks. 

\section{Runner Performance Pipeline}
\label{sec:pipeline}

\begin{figure}[t]  
% \begin{minipage}[t]{1\linewidth}
    \centering
    \includegraphics[scale=0.5]{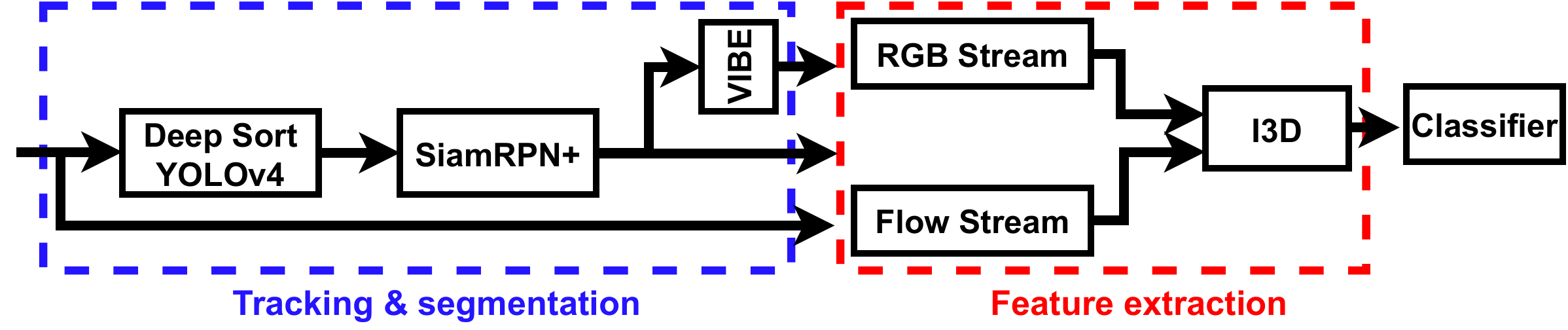}
    %   \end{minipage}
    \caption{\textbf{The proposed pipeline for the performance quality assessment context-driven problem.} The devised process comprises two major parts: the tracking and segmentation block and the features extraction block.}
 \label{fig:pipeline}
\end{figure}

As can be seen in Figure \ref{fig:pipeline}, this work proposes and evaluates a sequential pipeline divided into two major blocks. The first block is the subjects tracking and segmentation, and it provides the necessary information to locate and label the runner of interest in the scene. The output of this block can be divided into raw data, the runner bounding box (BB), and a fine-grained runner segmentation using the Video Inference for Human Body Pose and Shape Estimation (VIBE) \cite{kocabas19} (see Figure \ref{fig:context}). VIBE is a video pose and shape estimation method. It predicts the parameters of SMPL body model \cite{Loper2015} for each frame of an input video. Figure \ref{fig:pipeline} also shows how the tracking and segmentation block feeds the features extraction block. In more detail, the subject is located and given the same id (the same id indicates the same subject across frames) by the Deep SORT algorithm~\cite{Wojke17}. At the same time, the SiamRPN+ network~\cite{zhang2019deeper} stabilizes this process by avoiding flickering detections among consecutive frames. The SiamRPN+ network ensures a proper segmentation to adjust the tracking process in case of Deep SORT failure. Trackers output also feeds the VIBE algorithm to obtain a more accurate segmentation of the runner-of-interest.

\subsection{Runners Tracking and Segmentation}
\label{sec:tracker}

In the past few years, the Simple Online and Realtime Tracking (SORT) \cite{Bewley16} has shown a remarkable performance in object tracking. Moreover, SORT with deep association metric (Deep SORT) \cite{Wojke17} has been proposed for pedestrian detection as an extension of the SORT algorithm. Deep SORT aims to track people and correctly label the subjects in the scene. Recently, Deep SORT has reported stable tracking results in the sporting context \cite{Host20}. Even though Deep SORT achieves overall good performance in tracking precision and accuracy, we have observed that illumination changes or partial occlusions can generate some tracking failures, such as detection flickering. In order to keep the Deep SORT label consistency, we have included a second tracker. The SiamRPN+ \cite{zhang2019deeper} plays a backup role for the Deep SORT algorithm. This neural network has been introduced as an evolution of SiamRPN. The SiamRPN+  benefits from a deeper backbone like ResNet, leading to a remarkable robustness \cite{Huang20}. 

\begin{figure}[t]  
\begin{minipage}[t]{1\linewidth}
    \centering
    \includegraphics[scale=0.5]{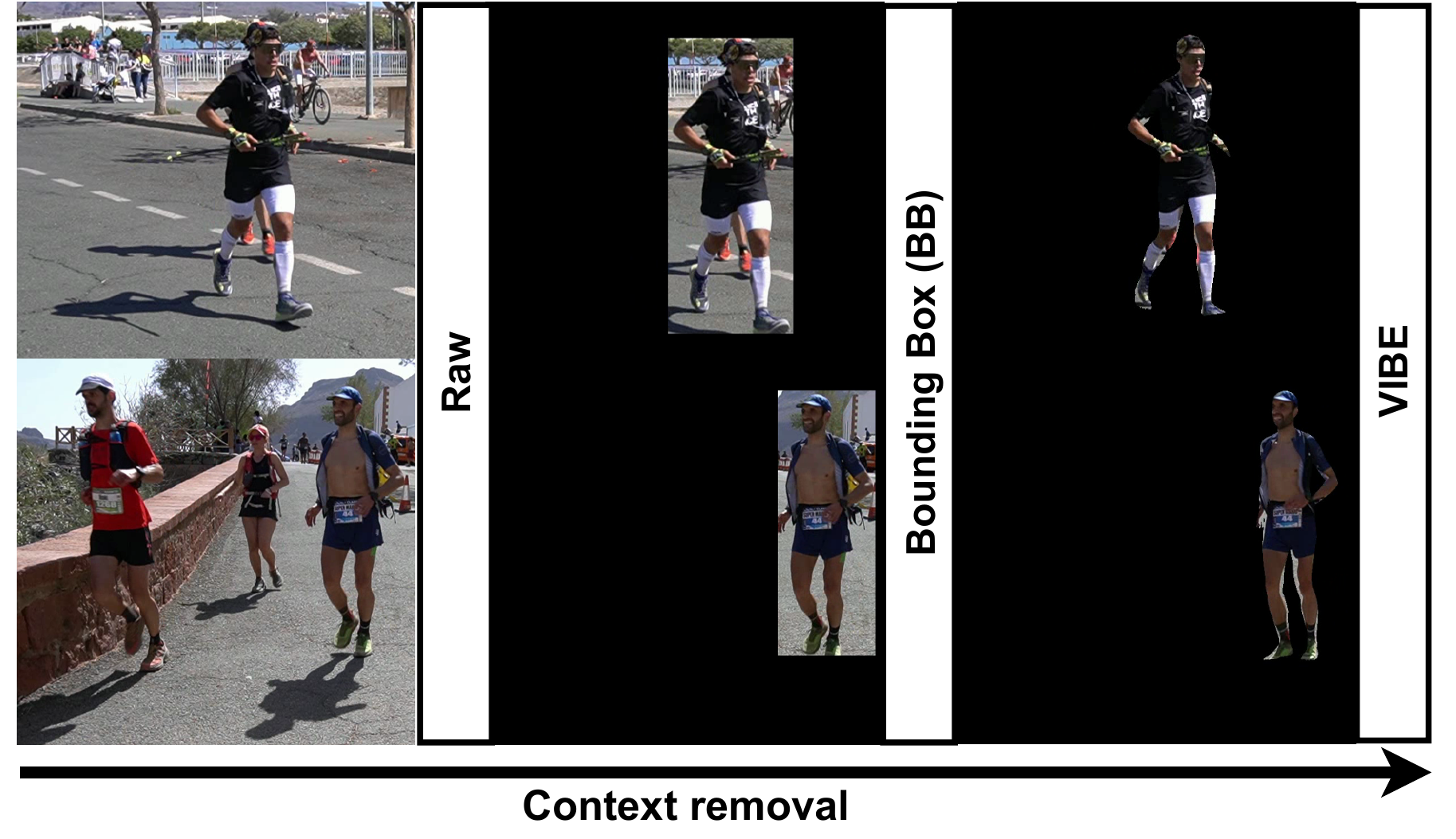}
      \end{minipage}
    \caption{\textbf{Levels of context removal.} We analyze the runner's performance with a model that efficiently classifies the quality assessment in a single forward pass and includes two types of context-removal processes. The first column shows original frames, while the second and third columns show frames where an increasing segmentation process is imposed.
     \label{fig:context}}
\end{figure}

Consequently, the previous described trackers provide a robust runner-of-interest bounding box (see Figure \ref{fig:context}, second column). Finally, VIBE makes use of the already detected bounding boxes to perform a runner fine-grain segmentation (see Figure \ref{fig:context}, third column). Then, several RNNs consider these features as input to process the sequential nature of human motion. Finally, a temporal encoder and regressor are used to predict the body parameters for the whole input sequence.
%Vibe fails in heavy occlusion, fast motion, and multi-person occlusion...de ahí que usemos las BB

\subsection{Runners Features Extraction}
\label{sec:i3d}

A few years ago, Carreira and Zisserman proposed the Inflated 3D Convnet (I3D) based on a two-stream network \cite{Carreira17}. This deep neural network applies a two-stream structure for RGB and optical flow to the Inception-v1 \cite{Szegedy15} along with 3D CNNs. Nowadays, I3D is one of the most common feature extraction methods for video processing. The approach presented in this work exploits the pre-trained model on the Kinetics dataset as a backbone model \cite{Carreira17}. Kinetics \cite{Kay17} is a large HAR dataset that includes a considerable number of action categories. Our proposal considered the backbone model trained on the Kinetics version of $400$ action categories.

Consequently, the I3D acts as a feature extractor to encode the network input into a $400$ vector feature representation that feeds the classifier.  Here, the $400$ output logits of the I3D are used as our classifier input. In this regard, five different classifiers were tested during the conducted experiments: Decision Tree, Random Forest, XGBoost, Linear SVM, and Logistic Regression. However, only the best classifier (XGBoost) results are reported in the next section.

\section{Experimental Evaluation}
\label{sec:results}

%This section is divided into two subsections related to the setup and results of the designed experiments, respectively. The first subsection describes the used dataset for the proposed problem and technical details, such as the applied data split and the considered quality assessment metric for the runner performance evaluation. The achieved results are summarized in the second subsection.

\subsection{Experimental Setup}
\label{sec:expsetup}

\subsubsection{Dataset}
\label{subsec:dataset}

To evaluate below the ultra-runners performance variation, we have partially used the dataset published by Penate et al.~\cite{Penate20-prl}. The mentioned dataset was collected during an ultra-running competition, known as Transgrancanaria (TGC), held in March 2020. TGC comprises six running distances, but the annotated data covers just participants in the TGC Classic who must cover 128 kilometers in 30 hours, at most.

Although the TGC dataset contains annotations for almost 600 participants in six different RPs, just 214 of them were captured after km 84 with daylight, see Table~\ref{tab:recording-points}.
In our work, just the last three RPs are considered in the experiments below, when performance drops are more likely due to fatigue. Moreover, during the last RPs, the different performances among participants, the gap between leaders and last runners increases along the track. Therefore, the split time variance is higher, and performance can be analyzed more accurately.
For each participant, seven seconds clips at 25 fps are fed to the tracking block described in Section \ref{sec:pipeline}.

\begin{table}[t]
\caption{RPs locations, times and kilometer (extracted from~\cite{Penate20-prl}). In this work we have considered the locations in bold type.}%onf stands for original number of frames, and
\label{tab:recording-points}
\centering
\begin{tabular}{c|c|c|c|c}
 Location & Km & Start Rec. Time & Footage (frames) & \# annot. runners \\ \hline
%RP1    & no & 1& 29760 & - & -\\ %Salida
RP1 & 16.5 & 00:06 & 140,616 & 419 \\     %Arucas Santidad
RP2 & 27.9 & 01:08 & 432,624 &586 \\       %Teror
\textbf{RP3} & \textbf{84.2} & \textbf{07:50} & \textbf{667,872} & \textbf{203} \\   %Presa
\textbf{RP4} & \textbf{110.5} & \textbf{10:20} & \textbf{1,001,208} & \textbf{139} \\   %Ayagaures
\textbf{RP5} &  \textbf{124.5} & \textbf{11:20} & \textbf{1,462,056} & \textbf{114} \\ \hline %Parque
% & 3734136 \\
\end{tabular}%
\end{table}

\subsubsection{Quality Assessment Metric}
\label{subsec:metric}

The quality of the observed movement is assessed considering the runner's RP split time. The used dataset contains runners footage at $k$ different RPs where $n^p$ are the runner's samples at the point $p$.
Then, there is a footage ($v_i^p$) for each runner ($i$) that represents his/her passing through the RP ($p$), where $t_i^p$ stands for the runner split time. 

In this work, the runner-performance estimation problem is modeled as a classification problem rather than a regression problem. Therefore, each runner must be associated with a category. Let $d$ be the function that maps the time of each runner into a category, then $c_i^p=d(t_i^p)$ represents the category associated with runner $i$ at a given RP ($p$). As aforementioned, we are using a set of descriptors ($\mathbf{x}_i^p$) associated with each footage (see Section \ref{sec:i3d}). Consequently, for each RP a dataset is defined as $D^p=\{(\mathbf{x}_i^p,c_i^p)\}$ and the complete dataset can be formulated as $D=D^1 \cup \dots \cup D^k$.

We previously stated that this work aims to check whether it is possible to estimate a runner's performance or not by evaluating just a few seconds of footage samples from a running that may last from 13 hours for the winner and 30 hours for the last runner. Furthermore, this estimation can be computed at the same RP, or at a later one is given the footage-motion data ($v_i^p$). To achieve this task, our purpose is to find a function $f$ such that $\hat{c}_i^{p+j}= f(\mathbf{x}_i^p)$ for $j=0 \dots k-p$.

The footage descriptors $\mathbf{x}_i^p$ are computed by the pre-trained I3D ConvNet. An equal-width discretization strategy is considered as a mapping function $d$ between the split time and the different categories. For example, in Table 2, when two categories are considered,  category one corresponds to those runners with a split time lower than the median, and category two corresponds to those with split time above the median. Similarly, for four categories, the runners are labeled as category one, two, three, and four as those whose split times are in the first, second, third, and fourth quartile, respectively. Finally, the runner-performance estimation is considered at the same RP ($\hat{c}_i^p= f(\mathbf{x}_i^p)$), and in the next RP  ($\hat{c}_i^{p+1}= f(\mathbf{x}_i^p)$). The results presented in the next section refer to the average accuracy computed on 100 iterations. For each iteration, train and test data are chosen randomly, and the results are averaged after considering a stratified 4-fold cross-validation.

\begin{table}[t]
\centering
\caption{Accuracy achieved for the current (Curr) and the next RP estimation. Each row considers a different number of categories. The last column shows the results considering a 3D convolution network instead of an I3D ConvNet.}
\label{table_rp}
\begin{tabular}{*4c|c|c}
\toprule
RP-\#  Categ. &  Raw & BB & VIBE & C3D\cite{Varol18} & 3DRes\cite{Hara18}\\
\midrule
Curr-$2$ & $\mathbf{83.7 \pm 2.8}$\% & $77.5 \pm 2.8$\% & $76.1 \pm 3.1$\% & $74.3 \pm 4.5$\% & $81.3 \pm 2.9$\%  \\
Curr-$3$ & $62.2 \pm 3.4$\% & $58.9 \pm 4.1$\% & $57.2 \pm 3.8$\% & $53.9 \pm 3.8$\% & $61.5 \pm 3.6$\%\\
Curr-$4$ & $54.4 \pm 3.6$\% & $45.4 \pm 3.6$\% & $46.6 \pm 3.8$\% & $37.5 \pm 2.9$\% & $47.1 \pm 3.7$\%\\ \midrule
Next-$2$ & $\mathbf{77.1 \pm 2.4}$\% & $73.5 \pm 3.0$\% & $73.9 \pm 2.5$\% & $72.1 \pm 4.9$\% & $75.5 \pm 2.9$\%\\
Next-$3$ & $59.7 \pm 3.3$\% & $55.3 \pm 3.8$\% & $53.8 \pm 3.5$\% & $52.5 \pm 3.4$\% & $57.7 \pm 2.4$\%\\
Next-$4$ & $49.1 \pm 3.5$\% & $42.5 \pm 3.1$\% & $40.0 \pm 3.6$\% & $36.5 \pm 2.7$\% & $46.3 \pm 3.8$\%\\
\bottomrule
% \bigskip
\end{tabular}
%\caption{Accuracy achieved for the next RP estimation.}
%\label{table_next_rp}
%\begin{tabular}{*4c|c}
%\toprule
%\# Categories &  Raw & BB & VIBE & C3D\\
%\midrule
%$2$ & $\bm{77,1 \pm 2,4}$\% & $\bm{73,5 \pm 3,0}$\% & $73,9 \pm 2,5$\% & $72,1 \pm 4,9$\%\\
%$3$ & $59,7 \pm 3,3$\% & $55,3 \pm 3,8$\% & $53,8 \pm 3,5$\% & $52,5 \pm 3,4$\%\\
%$4$ & $49,1 \pm 3,5$\% & $42,5 \pm 3,1$\% & $40,0 \pm 3,6$\% & $37,5 \pm 2,7$\%\\
%\bottomrule
%\end{tabular}
\end{table}

\subsection{Experimental Results}
\label{sec:expeval}

We conducted a set of experiments to validate the effectiveness of the described proposal. These experiments took place through a grid search considering different classifiers (see Section \ref{sec:i3d}). Only the best classifier (XGBoost) results are reported in this section, with a configuration of 200 estimators, a maximum tree depth of seven, and cross-entropy loss. The most relevant parameters were the number of categories, the I3D ConvNet input, and the classifier configuration.

As we argued in Section \ref{subsec:metric}, the number of categories can be fixed, providing different perspectives over the data. Using a lower number of categories where the density of the underlying classes is high (i.e., two categories for good and bad performance) maximizes the number of available samples per class. Alternatively, using a higher number of categories reduces the number of available samples per class. The second relevant parameter related to our proposal is the I3D ConvNet input. In this sense, we have described three possible configurations in Section \ref{sec:tracker}: original video sequence (Raw), the runner-of-interest bounding boxes (BB), and the runner-of-interest fine-grained segmentation (VIBE). 

In this work, the performance of the runners has been divided into different categories. Good/Poor performance corresponds to over/under the median split time. Good/Average/Poor corresponds with the first/second/third tertiles. Excellent/Good/Poor/Bad corresponds to the quartile of the split time.

Table \ref{table_rp} is divided into two horizontal blocks. The first block shows the accuracy of the quality assessment for a runner at a given RP. It means that our system can predict the runner performance using the embeddings extracted from the RP video sequence as input. Results are pretty promising when considering two categories the raw input reported rates are around $84$\%. Observe that this classification is done using only seven seconds captured from RPs located more than 10 km away from each other. The classifier also achieves noticeable rates when BB (around $78$\%) and VIBE (around $76$\%) configurations feed the I3D ConvNet. As expected, performance drops as the number of categories increases. It happens because of the classes redistribution.  In other words, having the same number of samples divided into a higher number of categories affects the model.

\begin{figure*}[t]  
\begin{minipage}[t]{\linewidth}
    \centering
    \includegraphics[scale=0.35]{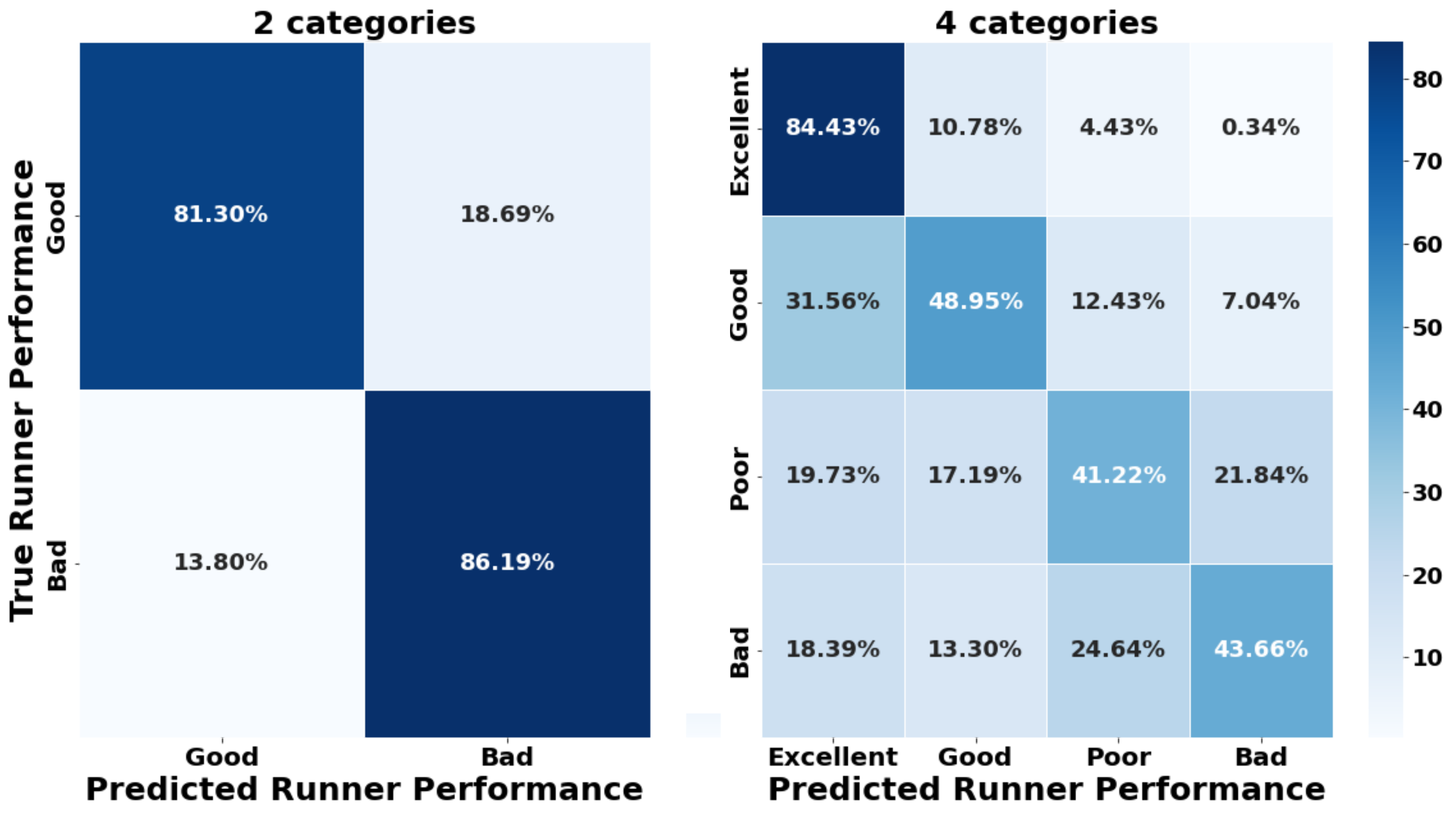}
      \end{minipage}
    \caption{\textbf{Current RP confusion matrices considering raw data as input.}
     \label{fig:conf}}
\end{figure*}

Table \ref{table_rp} second block shows the results of the estimated quality assessment for a runner on the next RP. Results are slightly worse than the rate reported in the first block of the table. It makes sense because the model predicts how the participant will perform on the next RP by observing the current RP. However, the results are quite interesting. Between the two experiments, the performance drops around $3-5$\%. At the same time, some runners tend to improve performance when they realize that they are being recorded (See Figure \ref{fig:conf}). Table \ref{table_rp} shows how the model performs on different inputs under the number of categories variations, i.e., how the context reduction affects the model performance. Three interesting issues must be highlighted in this regard.

%Moreover, the confusion matrices suggest more robustness among high-performance runners.  tienes razon, al no tener matriz de confusion de la segunda, mejor quitarla.

\begin{figure*}[t]  
% \begin{minipage}[t]{1\linewidth}
    \centering
    \includegraphics[scale=0.45]{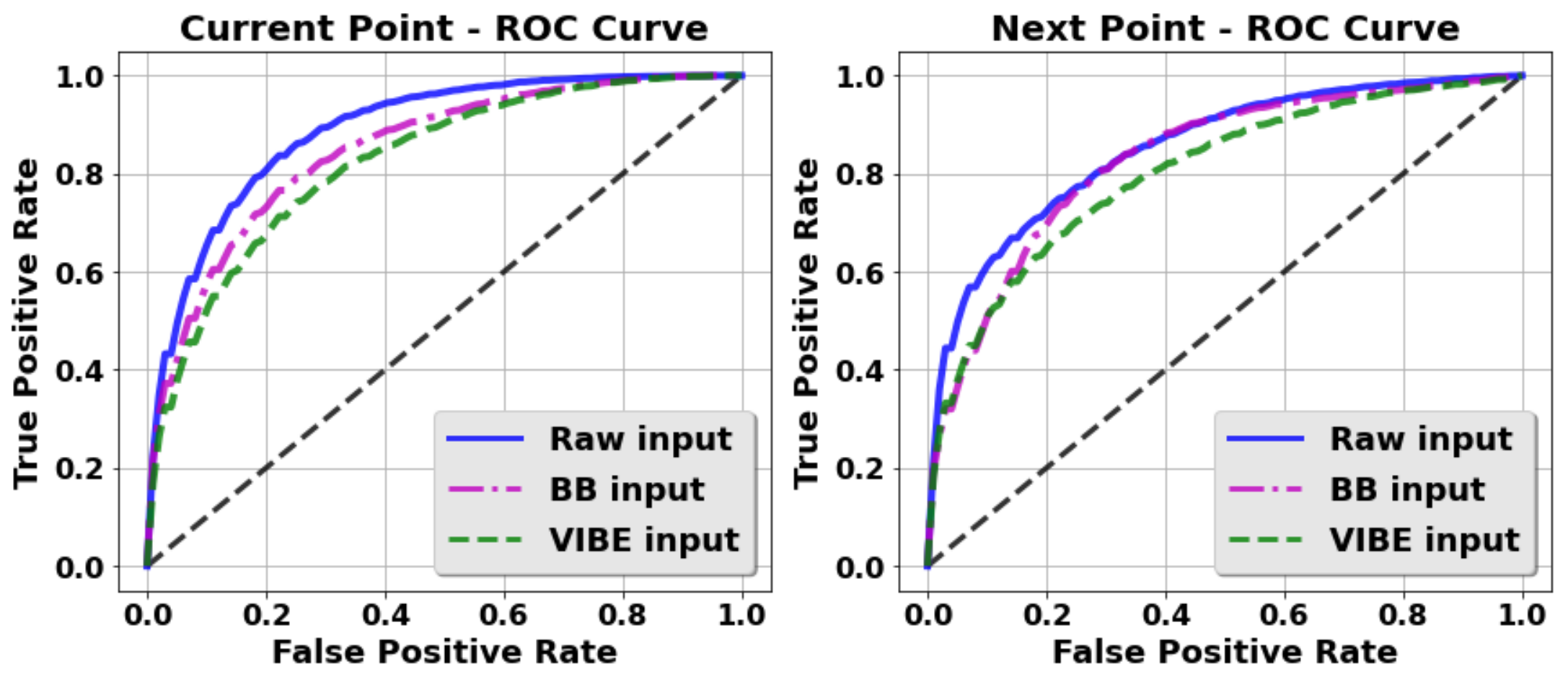}
    %   \end{minipage}
    \caption{\textbf{ROC curves computed from the results of the devised experiments when two categories are considered.} The left graph shows the ROC curves for the current RP experiment, while the right graph shows the ROC curves for the next RP experiment.
     \label{fig:roc}}
\end{figure*}

First, when the raw video sequence is considered as input, the model performs best in any considered case. It can be explained because the I3D ConvNet has been pre-trained using the Kinetics 400 (see Section \ref{sec:i3d} for a further description). This dataset provides context to the model, meaning that contextual information plays a relevant role. Moreover, the I3D ConvNet may experience difficulties through the optical flow stream input when the context is removed. In this sense, we believe that the feature matching could fail for regions without context. The table shows a $3-5$\% accuracy loss between the Raw and the BB inputs. This loss is lower between the BB and the VIBE inputs (around $1-2$\%) because the amount of context information removed is not much (see Figure \ref{fig:context}). It seems to happen that the model is not affected by multiple runners in the scene in order to evaluate a specific runner performance. The second issue is related to the best classifier. As aforementioned, XGBoost reports the best rates. It can be explained by observing how this algorithm works. While traditional Random Forest builds trees in parallel, in boosting, trees are built sequentially, i.e., each tree is grown and boosted using information from previously grown trees. Reducing the context may come with a computational advantage, i.e., the system may be faster if it only needs to process a small fraction of the scenes \cite{Adhikari18}. Finally, we have considered a C3D and a 3D ResNet for comparison purposes (see Section \ref{sec:sota}). Contrary to the two streams of data (RGB and Flow) used I3D ConvNet,  both C3D and 3D ResNets operate on single 3D stream input. As can be seen in Table \ref{table_rp}, I3D ConvNet outperforms C3D and 3D ResNets by a 10\%-15\% and a 2\%-5\% respectively.

\section{Conclusions}
\label{sec:con}

This paper presents an HAE approach to automatically provide an assessment for running quality and provide interpretable feedback. For this reason, we have conducted several experiments that combine an HAR pre-trained deep neural network with a quality assessment metric. The contribution represents an exciting challenge since we are unaware of any HAE research on this scenario. 

We have proven that both human motion and environmental backgrounds facilitate the necessary spatio-temporal information to the I3D ConvNet to generate a set of valid embeddings. Contrary to the works detailed in Section \ref{sec:sota}, the quality assessment metric is not defined by a collection of body configurations (i.e., as happens in diving or skating). The assessment metric relies on categorizing the runner's split time at each RP. Consequently, our proposal is sufficiently complex to effectively distinguish between the previously categorized classes with $83.7\%$ of accuracy at the current RP. We have also discussed how varying categories can affect the model accuracy. In this sense, scalability has been evaluated, and increasing the number of categories does not seem to improve a simpler model. This effect is caused by reducing samples per class during the training stage. Finally, we have shown that our proposal can predict the runner performance at the next RP with a $77.1\%$ of accuracy. Of course, health surveillance must be mentioned among the most relevant uses. Athletes face physically demanding situations in ultra-distance races. When they reach RPs, the race staff has first-hand information about a runner's state by watching him. For the race organizers, it is more feasible to place a set of cameras than placing medical staff along the different race tracks due to human resources. Our proposed technique provides a categorization that can be informative for the medical staff to evaluate runners' risky situations at any given time.

% esto iba antes de as a consequence
%

\bibliographystyle{splncs04}

%Author, F.: Article title. Journal \textbf{2}(5), 99--110 (2016)

%Author, F., Author, S.: Title of a proceedings paper. In: Editor,
%F., Editor, S. (eds.) CONFERENCE 2016, LNCS, vol. 9999, pp. 1--13.
%Springer, Heidelberg (2016). \doi{10.10007/1234567890}

%Author, F., Author, S., Author, T.: Book title. 2nd edn. Publisher,

%Author, A.-B.: Contribution title. In: 9th International Proceedings
%on Proceedings, pp. 1--2. Publisher, Location (2010)

%LNCS Homepage, \url{http://www.springer.com/lncs}. Last accessed 4
%\end{thebibliography}
\end{document}